\documentclass[runningheads]{llncs}

\usepackage{pifont}
\usepackage[utf8]{inputenc}
\usepackage{marvosym} 

\usepackage{graphicx}
\usepackage[T1]{fontenc}
\usepackage{CJKutf8}  
\usepackage{multirow}
\usepackage[table]{xcolor}
\usepackage{colortbl}
\definecolor{myblue}{RGB}{240,240,255}
\definecolor{dblue}{RGB}{180,180,245}
\usepackage[normalem]{ulem}
\definecolor{lgreen}{RGB}{236, 255, 220}
\definecolor{lred}{RGB}{255, 240, 255}
\definecolor{softgreen}{rgb}{0.4, 0.7, 0.3}
\definecolor{green}{RGB}{0, 255, 0}
\definecolor{gainlow}{RGB}{255,255,255}    % 白色
\definecolor{gainhigh}{RGB}{255,220,200}
\usepackage{graphicx}
\usepackage{booktabs}
\usepackage{amsmath} 
\usepackage{changepage}
\usepackage{arydshln}
\usepackage{float}
\usepackage{placeins}

\usepackage[symbol*]{footmisc} % 使用 symbol* 选项自定义脚注符号
\DefineFNsymbols*{custom}{{\dag}{\ddag}{\S}{\P}}

\usepackage{hyperref}
\usepackage{tabularx}
\usepackage{threeparttable}

\usepackage[T1]{fontenc}

\newcommand{\mypara}[1]{\vspace{.05in}\noindent\textbf{#1}\quad}

\begin{document}
\title{HiM2SAM: Enhancing SAM2 with Hierarchical Motion Estimation and Memory Optimization \\towards Long-term Tracking}
\titlerunning{Enhancing SAM2 with Motion and Memory Optimization}
\author{
  Ruixiang Chen\inst{1}\thanks{This work was partially conducted at IIS, ETH Zurich.} \and
  Guolei Sun\inst{2}\thanks{Corresponding author: Guolei Sun.} \and
  Yawei Li\inst{3} \and
  Jie Qin\inst{4} \and
  Luca Benini\inst{3}
}%
\authorrunning{R. Chen et al.} 
\institute{
  KTH Royal Institute of Technology, Stockholm, Sweden \and College of Computer Science, Nankai University, Tianjin, China \and IIS, ETH Zurich, Zurich, Switzerland \and
  Nanjing University of Aeronautics and Astronautics, Nanjing, China
   \\
  \email{louischencrx@gmail.com, guolei.sun@nankai.edu.cn, yawli@iis.ee.ethz.ch, jie.qin@nuaa.edu.cn, lbenini@iis.ee.ethz.ch}
}

\maketitle              % typeset the header of the contribution
\begin{abstract}
This paper presents enhancements to the SAM2 framework for video object tracking task, addressing challenges such as occlusions, background clutter, and target reappearance. We introduce a hierarchical motion estimation strategy, combining lightweight linear prediction with selective non-linear refinement to improve tracking accuracy without requiring additional training. In addition, we optimize the memory bank by distinguishing long-term and short-term memory frames, enabling more reliable tracking under long-term occlusions and appearance changes. Experimental results show consistent improvements across different model scales. Our method achieves state-of-the-art performance on LaSOT and LaSOT$_\text{ext}$ with the large model,  achieving 9.6\% and 7.2\% relative improvements in AUC over the original SAM2, and demonstrates even larger relative gains on smaller models, highlighting the effectiveness of our trainless, low-overhead improvements for boosting long-term tracking performance. The code is available at \href{https://github.com/LouisFinner/HiM2SAM}{https://github.com/LouisFinner/\\HiM2SAM}.

\end{abstract}

\section{Introduction}

Visual object tracking (VOT) in open-world environments remains a fundamental yet challenging problem, particularly in scenarios involving complex motion, target disappearance and reappearance, and long-term tracking. Traditional tracking-by-detection methods (e.g.,~\cite{adpt,boostrack,ucmc}) often rely on predefined object classes and motion models such as linear Kalman filters~\cite{kf}, limiting their adaptability to the diverse and complex dynamics of real-world scenarios.

Recent class-agnostic tracking-by-propagation models, such as the Segment Anything Model 2 (SAM2)~\cite{ravi2024sam2}, offer improved generalization by leveraging memory-based architectures to segment targets across frames without retraining. While SAM2 performs well in standard scenarios, it is prone to failures in long-term tracking due to its reliance on visual similarity and limited memory capacity, especially under occlusion, fast motion, or the presence of distractors. These limitations, along with the lack of explicit motion modeling, often lead to identity switches and missed detections when visual cues are unreliable.

To address these issues, several methods enhance SAM2 via training-free modifications, such as SAMURAI’s~\cite{yang2024samurai} motion modeling with Kalman filters, SAM2Long’s~\cite{ding2024sam2long} confidence-based memory decisions, DAM4SAM’s~\cite{dam4sam} distractor-aware memory filtering, SAM2MOT's~\cite{jiang2025sam2mot} multi-object memory management. 

However, these efforts mostly improve isolated components of the SAM2 framework, either by introducing basic motion models or by designing memory mechanisms that handle specific cases like distractors. Most lack accurate motion estimation, which weakens temporal consistency. Moreover, memory strategies are often designed for narrow use cases and struggle to generalize to the diverse conditions of long-term tracking, particularly in scenarios involving complex motion and occlusion. These limitations highlight the need for more advanced motion modeling and a unified optimization of motion and memory to better address the diverse challenges of long-term tracking.

In this work, we introduce \textbf{HiM2SAM}—an enhanced SAM2 framework that tackles long-term and complex tracking challenges through two key innovations.
\textbf{1) Hierarchical Motion Estimation}: We combine lightweight Kalman-based linear motion prediction with pixel-level prediction applied only to selected frames via a point tracker, enabling the model to better handle complex object motion while maintaining low runtime overhead.
\textbf{2) Optimized Memory Structure}: We divide the memory bank into short-term and long-term components,  incorporating motion-aware filtering to selectively store high-confidence and distinctive frames. This design not only improves the reliability of memory representations but also enhances robustness against distractors and occlusions under limited memory capacity.

\begin{figure}[t]
    \centering
    \includegraphics[width=1.0\linewidth]{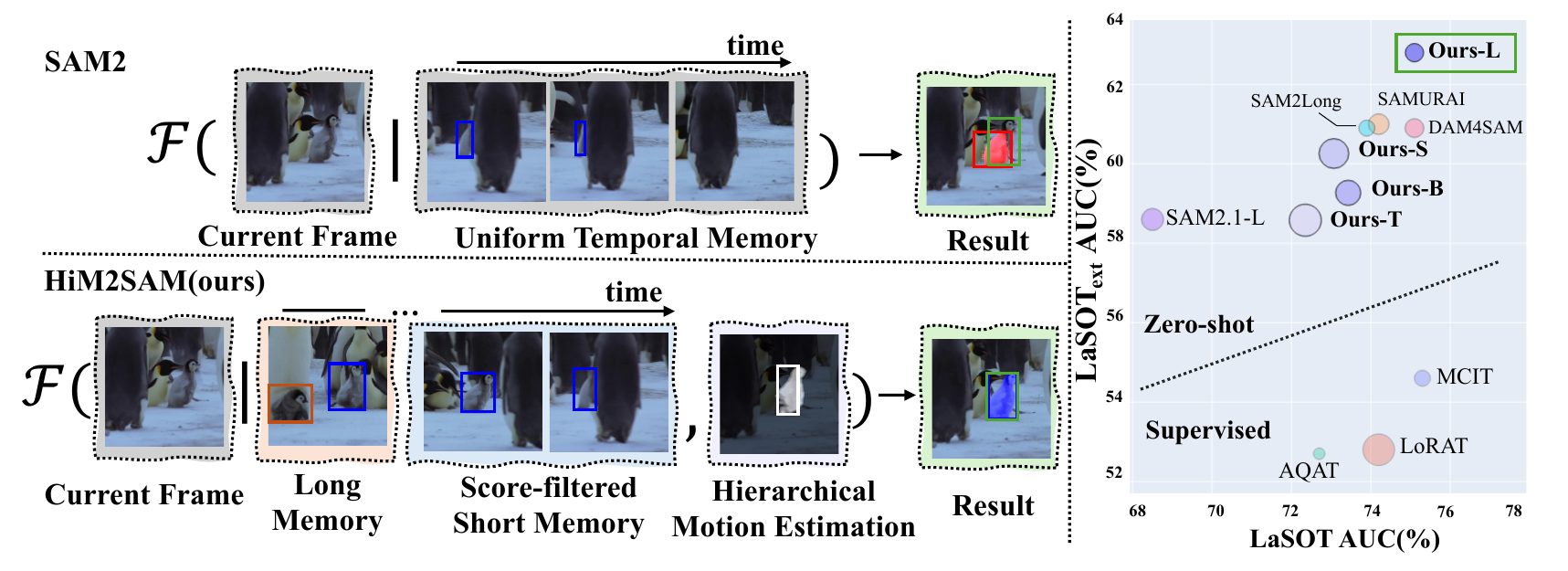}
    \caption{HiM2SAM with hierarchical motion estimation and long-short memory. \textbf{Left:}  \textcolor{blue}{Blue box}: bounding box of detected target object; \textcolor{orange}{Orange box}: bounding box of distractive object; \textcolor{green}{Green box}: ground truth; \textcolor{red}{Red box}: inaccurate tracking result; White box: coarse motion estimation; White mask: fine motion estimation. \textbf{Right:} Comparison with state-of-the-art trackers on LaSOT series datasets. Circle size indicates FPS; axes show AUC performance; larger is better for both. Our method includes four models of increasing size: T, S, B, and L.}
    \label{fig:maininfo}
\end{figure}

HiM2SAM remains training-free, incurs minimal additional cost, and delivers performance gains across VOT benchmarks, particularly on LaSOT~\cite{fan2020lasothighqualitylargescalesingle,fan2019lasothighqualitybenchmarklargescale} series datasets. It also significantly boosts performance on smaller SAM2 variants, highlighting its effectiveness in efficient, long-term object tracking. An overview of our method and its comparison with SOTA performance on LaSOT are shown in Figure~\ref{fig:maininfo}.

\section{Related Work}

\mypara{Transformer-based VOT.}
Transformer~\cite{vaswani2023attentionneed} architectures have achieved state-of-the-art (SOTA) results in various vision tasks including object tracking~\cite{zeng2021motr,cheng2022xmem} and segmentation~\cite{zhu2024clip,sun2022coarse}. Methods like MOTR~\cite{zeng2021motr} unify detection and tracking in a single end-to-end framework, while others such as SAMBA~\cite{segu2024samba} use Mamba-based state transitions to update object tokens over time. However, these recursive updates can lead to drift and weaken long-term temporal consistency. More recent approaches, inspired by the human visual system, shift toward segmentation-based tracking. Models like XMem~\cite{cheng2022xmem} and CUTIE~\cite{cheng2023putting} build memory banks using past image and mask features, enabling robust temporal object association through cross-attention mechanisms.

\mypara{Segment Anything Model 2 (SAM2).}
SAM2~\cite{ravi2024sam2} extends the Segment Anything Model (SAM)~\cite{kirillov2023segment} to video by introducing a memory bank that stores previous frame features and masks. During inference, cross-attention is computed between the current frame and memory frames to produce segmentation masks without requiring prompts. Its capabilities have been validated in various tasks~\cite{zhou2025sam2,cuttano2025samwise,li2025samjam}. Despite strong performance, SAM2’s limited memory capacity and lack of temporal fusion hinder its robustness against distractors and long-term occlusions. To mitigate this, SAMURAI~\cite{yang2024samurai} adds motion prediction and mask reweighting using Kalman filters; SAM2Long~\cite{ding2024sam2long} treats mask selection as a sequential decision-making task; and DAM4SAM~\cite{dam4sam} identifies distractor frames and leverages them to guide long-term memory updates, retaining high-confidence features for improved tracking stability.

\mypara{Point Tracking.}
Tracking points across video frames supports fine-grained tasks like 3D reconstruction and motion analysis, focusing on pixel-level correspondence rather than object semantics. Recent works such as  CoTracker~\cite{karaev2023cotracker} introduce differentiable correlation, global matching, and attention-based mechanisms to improve robustness. CoTracker3~\cite{karaev2024cotracker3} simplifies this paradigm by combining efficient 4D correlation and iterative track updates, achieving SOTA accuracy with significantly less training data. Its ability to model dense, long-term motion complements segmentation-based approaches, making it especially useful in complex, dynamic scenes.
% PIPs~\cite{harley2022particle}, 
% , and
% TAPIR~\cite{doersch2023tapir}

\section{Revisit Segment Anything Model 2}
\begin{figure}[t]
    \centering
    \includegraphics[width=0.95\linewidth]{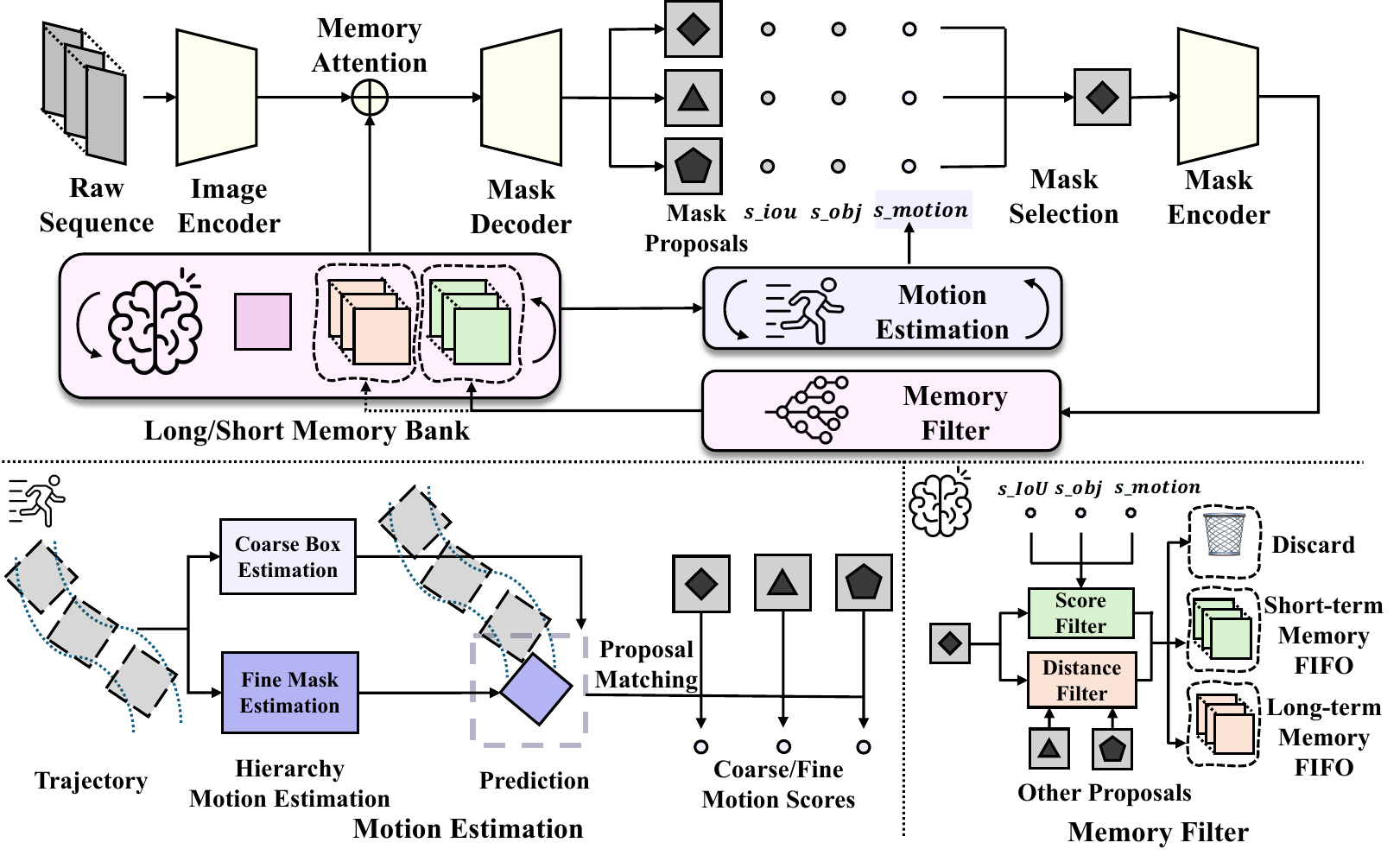}
    \caption{The structure of HiM2SAM. The upper part illustrates the overall architecture, where the \colorbox[RGB]{255,255,241}{yellow} blocks represent the original SAM2 components, while the \colorbox[RGB]{252,240,254}{red} and \colorbox[RGB]{240,240,254}{purple} blocks indicate the modules introduced by HiM2SAM. The bottom-left shows the hierarchical motion estimation module, while the bottom-right details the memory filter module.}

    \label{fig:mainmethod}
\end{figure}

SAM2 builds upon the architecture of SAM (Segment Anything Model) and extends it to video object segmentation by introducing a memory mechanism, which enables prompt-free segmentation across frames. In addition to a Vision Transformer-based image encoder~\cite{ryali2023hiera} that extracts per-frame visual features, the memory mechanism introduces several key components tailored for video segmentation: a mask decoder, a memory encoder and memory bank, and memory attention. The structure of SAM2 is shown in upper part in Figure~\ref{fig:mainmethod}.

\mypara{Mask Decoder.} The mask decoder in SAM2 receives memory-conditioned image embeddings and encoded prompt tokens as inputs. It uses multiple heads to generate a set of predicted masks, each with an Intersection over Union (IoU) score and an objectness score to assess the confidence of the mask. The highest-scoring mask is selected as the final output. Currently, SAM2 outputs three proposal masks.

\mypara{Memory Encoder and Memory Bank.}
The memory encoder encodes the predicted mask along with its associated image feature to produce memory representations, which are stored in a first-in-first-out (FIFO) memory bank that retains outputs from the most recent frames for later retrieval. This enables SAM2 to build a memory of object features and mask information over time, providing better context for tracking and segmentation. However, since the memory bank has a limited capacity, it may discard useful historical information, which can pose challenges for long-term tracking.

\mypara{Memory Attention.} The memory encoder stores fused features and masks from recent frames in a memory bank. For each new frame, the model computes cross-attention between its features and the stored object representations, while the memory attention layer performs both self-attention and cross-attention. This mechanism allows the model to integrate temporal information and enhance tracking consistency across frames. Despite its strong performance, SAM2 may overlook temporal motion cues, leading to tracking failures in cases of fast movement or occlusion, especially over long durations.

\section{Proposed Method} 

To tackle the challenges SAM2 faces in long-term tracking, we introduce a hierarchical pixel-level motion estimation framework to improve proposal disambiguation under complex motion patterns. For memory management, we adopt a motion-aware confidence filtering strategy to select reliable short-term memory frames. We further enrich the long-term memory bank by identifying distinctive frames with potential distractor-induced ambiguities, enhancing target representation. These components work in concert to improve robustness against occlusion, background clutter, and reappearance, ultimately enhancing performance in long-term and challenging tracking scenarios.

\subsection{Hierarchical Motion Estimation}
In complex open-world scenarios, object-level Kalman filter tracking can suffer from inaccuracies due to the complexity of object motion, whereas point tracking models provide finer-grained pixel-level motion estimation that better handles non-linear and detailed movements, maintaining high-confidence tracking over time. However, state-of-the-art methods like CoTracker3 are computationally expensive, and applying such optimization to every frame significantly reduces tracking speed.

To balance efficiency and accuracy, we introduce a hierarchical motion estimation strategy. For most frames where object motion is minor and appearance remains stable, a lightweight motion estimator is used to refine SAM2’s mask predictions. However, in challenging frames with occlusion or background clutter, where the predictions of both SAM2 and the motion estimator tend to have low confidence, a more precise non-linear motion estimator is employed by leveraging a point tracker. This selective design enables accurate motion modeling while maintaining real-time performance.

\mypara{Coarse Motion Estimation.} During the forward tracking process, adjacent frames are similar, and motion and shape changes are relatively smooth. Therefore, we use the Kalman filter to estimate the target’s bounding box, including the target’s center position, velocity of position changes, and edge length ratio. During initialization, the Kalman filter takes predictions as input to update its parameters. In the prediction phase, SAM2 predicts multiple proposal masks, and the Kalman filter predicts the position and shape of the target bounding box based on previous motion estimates.

We calculate the IoU between the Kalman filter’s predicted box and each proposal box, resulting in a coarse motion confidence score $s_{\text{coarse}}$. This score is linearly combined with the SAM2 mask confidence $s_{\text{iou}}$ using a hyperparameter $\alpha$. We select the proposal with the highest $s_{\text{conf}}$ as the final output.

\begin{equation}
s_{\text{conf}} = \alpha \cdot s_{\text{coarse}} + (1-\alpha) \cdot s_{\text{iou}}.
\end{equation}

\mypara{Fine Motion Estimation.} Kalman filter  assumes linear motion, and in cases of fast or complex object motion and occlusion, coarse motion estimation may not be reliable. When $s_{\text{conf}}$ is below a threshold $\tau$, we apply fine motion estimation.

For each of the three mask proposals in the current frame, we apply farthest point sampling~\cite{fps} and propagate the points backward to historical frames using a point tracker. In each historical frame, these tracked points are used to reconstruct a softmask $M^{(t)}$ via radial basis function (RBF)~\cite{rbf}. 

\begin{equation}
M^{(t)}(x) = \sum_{i=1}^{N} \phi(\|x - x_i\|),
\end{equation}

\noindent where $\phi(\cdot)$ is a Gaussian kernel, $x_i$ are the tracked visible points, and  $N$ is the number of visible points. The reconstructed mask $M^{(t)}$ is then compared with the historical prediction mask $M^{(h)}$ using the Dice coefficient~\cite{dice} to evaluate their similarity and temporal consistency, resulting in the similarity score $s_{\text{fine}}$. For the final decision, the score is obtained by linearly combining $s_{\text{coarse}}$ and $s_{\text{iou}}$ using hyperparameters $\alpha$ and $\beta$.

\begin{equation}
s_{\text{conf}} = \alpha \cdot s_{\text{coarse}} + \beta \cdot s_{\text{fine}} + (1 - \alpha - \beta) \cdot s_{\text{iou}}.
\end{equation}

\subsection{Long/Short Memory Bank}
In long video tracking scenarios, objects may be occluded or undergo rapid motion, causing them to be invisible for several frames. Due to SAM2’s FIFO memory bank structure, most recent memory frames may not have reliable target object feature information, relying only on the prompt frame from the first frame of the video for tracking. When a target reappears, its position and shape typically change significantly, leading to possible loss or incorrect tracking. To address this, we propose a division of the memory bank into long-term and short-term components.

The short-term memory bank is updated in a FIFO manner to retain recent temporal information. Frames with high SAM2 IoU scores and high motion estimation scores are selected as high-confidence predictions, ensuring reliable mask quality and temporal consistency. For the long-term memory bank, we further select a subset of these high-confidence frames that potentially contain distractor-induced ambiguities, storing them to enhance robustness against occlusion, distractors, and target reappearance.

To identify such frames more precisely, we examine the proposal masks from SAM2. In most normal frames without confusion, the proposal masks are highly consistent and often nearly identical, differing only in subtle boundary details or exhibiting partial inclusion due to varying semantic granularity. In contrast, in ambiguous frames, the proposal masks tend to be mutually exclusive, indicating that the model may have detected different objects, often distractors, rather than variations of the same target. By marking these frames as distinctive and incorporating their mask results into the long-term memory, we enrich the target representation, help the model disambiguate confusing cases, and improve tracking robustness under challenging conditions.

For each frame prediction, we first assess the confidence score. For high-confidence frames, we evaluate the separation between proposals, using the directed Hausdorff distance~\cite{hsdf} between mask contours. Given masks \(M^*\) and \(M_i\), their contours \(\partial M^*\) and \(\partial M_i\) are extracted, and the directed Hausdorff distance is defined as:
\begin{equation} 
\mathcal{H}(M^*, M_i) =  \sup_{y \in \partial M_i} \inf_{x \in \partial M^*} \|y - x\| .
\end{equation}

Frames that exhibit significant confusion due to background clutter, characterized by a large Hausdorff distance between the selected mask and alternative proposals, are added to the long-term memory bank to support future predictions.

\section{Experiments}

\subsection{Benchmarks}
\mypara{LaSOT.}
LaSOT~\cite{fan2019lasothighqualitybenchmarklargescale} is a single-object tracking benchmark from the VOT series, comprising 280 testing sequences. Each video contains an average of 2,500 frames at 30 FPS, corresponding to an average duration of approximately 83 seconds, which is significantly longer than many other benchmarks. 
The dataset is particularly challenging due to its long sequence durations, high appearance variation, frequent occurrences of occlusion and background clutter.

\mypara{LaSOT$_\text{ext}$.}
LaSOT$_\text{ext}$~\cite{fan2020lasothighqualitylargescalesingle} is an extension of the LaSOT dataset, adding 150 new testing videos and 15 additional object categories. The video lengths and frame rates are consistent with LaSOT. Compared to LaSOT, LaSOT$_\text{ext}$ introduces a greater number of challenging cases, including small objects, fast-moving targets, and low-resolution videos, posing significant challenges for visual tracking.

\mypara{VOT-LT2020.} The VOT challenge is a widely recognized benchmark series for evaluating visual trackers under standardized protocols. VOT-LT2020~\cite{kristan2020eighth} introduces a long-term tracking benchmark consisting of 50 challenging sequences covering diverse object categories such as persons, vehicles, animals, and more, with a total of 215,294 frames. On average, each sequence includes 10 long-term target disappearances, each lasting approximately 52 frames. 

\mypara{VOT-LT2022.}VOT-LT2022~\cite{kristan2022tenth} further refines this setup by providing 50 carefully selected sequences with 168,282 frames in total at a fixed resolution of 1280×720, featuring frequent disappearances and prolonged occlusions to better evaluate the robustness of long-term tracking methods.

\begin{table}[t]
\centering
\caption[Tracker Performance Comparison]{Comparison of recent SOTA trackers on long-term LaSOT series datasets. \colorbox{myblue}{Purple} highlights our method; \textbf{bold} indicates the top result within each model size. T, S, B, and L denote Tiny, Small, Base, and Large configurations, respectively, with weights aligned to those of the original SAM2 models.}
\resizebox{0.95\textwidth}{!}{ 

\begin{tabularx}{\textwidth}{l*{6}{>{\centering\arraybackslash}X}}
\toprule
\label{tab:main_results}
\multirow{2}{*}{\textbf{Trackers}}
& \multicolumn{3}{c}{\textbf{LaSOT$_{\text{ext}} $}} & \multicolumn{3}{c}{\textbf{LaSOT}}  \\
\cmidrule(lr){2-4} \cmidrule(lr){5-7} 
& {{AUC(\%)}} & {{P$_{\text{norm}}$(\%)}} & {{P(\%)}}  & {{AUC(\%)}}  & {{P$_{\text{norm}}$(\%)}} & {{P(\%)}}  \\
\midrule
\textit{Supervised method}  &   &  &   &   & &   \\
EVPTrack-B \cite{shi2024evptrack} & 53.7 & 65.5 & 61.9 & 72.7 & 82.9 & 80.3 
\\
ODTrack-B \cite{zheng2024odtrack} & 52.4 & 63.9 & 60.1 & 73.2 & 83.2 & 80.6 
\\
HIPTrack-B \cite{cai2024hiptrackvisualtrackinghistorical} & 53.0 & 64.3 & 60.6  & 72.7 & 82.9 & 79.5
\\
AQATrack-B \cite{xie2024autoregressive}  & 51.2 & 62.2 & 58.9 & 71.4 & 81.9 & 78.6
\\
AQATrack-L & 52.7 & 64.2 & 60.8  & 72.7 & 82.9 & 80.2 
\\
LoRAT-B224 \cite{lorat} & 50.3 & 61.6 & 57.1 & 71.7 & 80.9 & 77.3 
\\
LoRAT-L224   & 52.8 & 64.7 & 60.0& 74.2 & 83.6 & 80.9
\\
MCITrack-B \cite{kang2025exploring} & 54.6 & 65.7 & 62.1  & 75.3 & 85.6 & 83.3
\\

\midrule
\textit{Zero-shot method}  &   &  &   &  & & \\
SAM2.1-T\cite{ravi2024sam2}  & 52.3 & 62.0 & 60.3& 66.7 & 73.7 & 71.2\\
SAMURAI-T\cite{yang2024samurai}   & 55.1  & 65.6 & 63.7 & 69.3 & 76.4 & 73.8\\
DAM4SAM-T\cite{dam4sam}  & 57.9 & 68.6 & 67.5 & 72.2 & 79.8 & 77.1 \\
\rowcolor{myblue}\textbf{HiM2SAM-T(Ours)} & \textbf{58.6}  & \textbf{69.3}  & \textbf{68.0}  & \textbf{72.4}  & \textbf{80.3}  & \textbf{78.0} 
\\
\hdashline

SAM2.1-S~  & 56.1 & 67.6 & 65.8 & 66.5 & 73.7 & 71.3\\
SAMURAI-S & 58.0  & 69.6 & 67.7  & 70.0  & 77.6  & 75.2\\
DAM4SAM-S & 53.7 & 60.8 & 60.1 & 66.9 & 69.2 & 66.0  \\

\rowcolor{myblue}\textbf{HiM2SAM-S(Ours)} & \textbf{60.3}  & \textbf{71.9}  & \textbf{ 70.4} & \textbf{73.1}  & \textbf{81.0}  & \textbf{78.8} 
\\
\hdashline
SAM2.1-B~ & 55.5 & 67.2 & 64.6 & 66.0 & 73.5 & 71.0  \\
SAMURAI-B & 57.5  & 69.3  & 67.1 & 70.7 & 78.7  & 76.2  \\
DAM4SAM-B & 58.6 & 69.4 & 68.2  & 73.3 & 81.3 & 78.8 \\

\rowcolor{myblue}\textbf{HiM2SAM-B(Ours)}  & \textbf{59.3}  & \textbf{71.2}  & \textbf{69.4}  & \textbf{73.4}  & \textbf{81.7}  & \textbf{79.5}
\\
\hdashline
SAM2.1-L~ & 58.6 & 71.1 & 68.8& 68.5 & 76.2 & 73.6  \\
SAM2Long\cite{ding2024sam2long}  & 60.9 & - & -  & 73.9 & - & - \\%& 81.1 & - & -
SAMURAI-L  & 61.0 & 73.9 & 72.2& 74.2 & 82.7  & 80.2 \\
DAM4SAM-L  & 60.9 & - & - & \textbf{75.1} & - & - \\
\rowcolor{myblue}\textbf{HiM2SAM-L(Ours)}  & \textbf{62.8}  & \textbf{75.5}  & \textbf{74.3} & \textbf{75.1}  & \textbf{83.2}  & \textbf{81.0} \\
\bottomrule
\end{tabularx}
}
\end{table}

\subsection{Quantitative Results}
\mypara{Overal Results.}
We compare state-of-the-art supervised tracking methods and zero-shot trackers. Unless otherwise specified, all SAM2-based experiments in this paper refer to the latest SAM2.1 model.  For our method, CoTracker3~\cite{karaev2024cotracker3} is employed for point propagation. To ensure fair comparison, only methods capable of achieving real-time performance are considered. Table~\ref{tab:main_results} presents the quantitative results on LaSOT series. Our proposed method achieves competitive overall performance among zero-shot methods. Despite the absence of training data, HiM2SAM demonstrates significant gains over supervised approaches on LaSOT$_\text{ext}$, showcasing the strong generalization ability. 
 
Furthermore, our method consistently outperforms baseline approaches across all SAM2 model sizes. The gains are especially notable in smaller models, which often struggle with complex tracking due to limited capacity. This demonstrates the value of our hierarchical motion estimation and long-short memory mechanism. By enhancing frame correspondence and managing memory efficiently, these components improve tracking robustness with minimal computational overhead, making them particularly beneficial for lightweight models. Importantly, these improvements are achieved without sacrificing the class-agnostic nature of SAM2, demonstrating that our approach maintains strong robustness and generalization across diverse objects and scenarios.

As shown in Table~\ref{tab:vot}, we evaluate zero-shot performance on other two long-term tracking benchmarks: VOT-LT2020 and VOT-LT2022 using the large-size models. Compared with previous zero-shot methods, HiM2SAM achieves the highest F-score on both datasets, indicating stronger robustness under long-term tracking scenarios. These results suggest that our design, which incorporates hierarchical motion estimation and memory management, is effective in handling frequent target disappearances and complex scene variations.

\begin{table}[t]
\centering
\small
\caption{Comparison of zero-shot methods on long-term tracking benchmarks VOT-LT2020 and VOT-LT2022 using the large-size models. \colorbox{myblue}{Purple} highlights our method; \textbf{bold} indicates the top result under zero-shot setting.}

\resizebox{0.95\textwidth}{!}{  

\begin{tabularx}{\textwidth}{l*{6}{>{\centering\arraybackslash}X}}
\toprule
\label{tab:vot}
\multirow{2}{*}{\textbf{{Trackers}}}
& \multicolumn{3}{c}{\textbf{{VOT-LT2020}}} & \multicolumn{3}{c}{\textbf{{VOT-LT2022}}}  \\
\cmidrule(lr){2-4} \cmidrule(lr){5-7} 
 & {{Pr}} & {{Re}}  & {{F-score}}  &  {{Pr}} & {{Re}}  & {{F-score}}\\
\midrule
\textit{Zero-shot method}  &   &  &   &  & & \\
SAM2.1~\cite{ravi2024sam2} & 0.602 & 0.689 & 0.643 & 0.428 & 0.485  & 0.455
 \\
SAMURAI~\cite{yang2024samurai} & 0.586 & 0.670 & 0.625 & 0.445 & 0.509 & 0.475
 \\
DAM4SAM~\cite{dam4sam} & 0.607 & 0.698 & 0.649
&0.436 & 0.499 & 0.465
  \\

\rowcolor{myblue}\textbf{HiM2SAM(Ours)}& \textbf{0.612}  & \textbf{0.703}  & \textbf{0.654}  & \textbf{0.450}  & \textbf{0.513}  & \textbf{0.480} 
\\

\bottomrule
\end{tabularx}
}
\end{table}

\mypara{Tracking Attributes.}
We further conduct an attribute-based analysis on both LaSOT and LaSOT$_\text{ext}$ using the base-size model, focusing on AUC performance. The attributes defined in~\cite{fan2019lasothighqualitybenchmarklargescale} can be categorized into two categories, Dynamic Motion and Degraded Visibility. As shown in Table~\ref{tab:attributes}, our method consistently outperforms SAMURAI across all attributes, indicating strong generalization to diverse and challenging tracking scenarios.

In the Dynamic Motion group, our method achieves substantial gains by introducing hierarchical motion estimation and motion-guided memory filtering, which enhance temporal consistency and motion robustness. Notably, significant improvements are observed in view change (+4.0 VC), camera motion (+3.0 CM), fast motion (+2.1 FM), and deformation (+2.1 DEF), highlighting the effectiveness of our design in handling complex motion and view variations.

In the Degraded Visibility group, HiM2SAM incorporates both motion-aware and distance-based memory filtering to better manage temporal information under challenging conditions. The motion filter enhances the reliability of memory frames by leveraging motion estimation, while the distance filter selectively preserves relevant long-term memory, mitigating the impact of occlusion and distractors. This design yields clear improvements in partial occlusion (+2.6 POC), background clutter (+2.4 BC), and full occlusion (+1.8 FOC). The method also shows consistent gains under low resolution (+2.3 LR) and illumination variation (+1.3 IV), demonstrating stronger resilience to degraded visibility and appearance ambiguity.

\begin{table}[t]
\centering
\caption{Attribute-wise AUC (\%) comparison between SAMURAI and HiM2SAM on LaSOT and LaSOT$_\text{ext}$ using base-size models. Attributes are grouped into {Dynamic Motion} and {Degraded Visibility}. The bottom row shows the average gains of our method over SAMURAI, with cell colors indicating the magnitude of improvement using a purple gradient.}
\label{tab:attributes}
\small
\begin{tabularx}{1.\textwidth}{l
*{14}{>{\centering\arraybackslash}X}
}
\toprule
\multirow{2}{*}{\textbf{Trackers}} & 
 \multicolumn{8}{c}{\textbf{{Dynamic Motion}}} & \multicolumn{6}{c}{\textbf{{Degraded Visibility}}} \\
\cmidrule(lr){2-9} \cmidrule(lr){10-15}
 &  
\scriptsize{VC} & \scriptsize{CM} & \scriptsize{DEF} & \scriptsize{ARC} & \scriptsize{FM} & \scriptsize{SV} & \scriptsize{ROT} & \scriptsize{MB} & 
\scriptsize{POC} & \scriptsize{BC} & \scriptsize{LR} & \scriptsize{FOC} & \scriptsize{IV} & \scriptsize{OV} 
\\
\midrule
 \multirow{1}{*}{\textbf{}} & \multicolumn{14}{c}{\textbf{LaSOT}} \\
\midrule
SAMURAI~\cite{yang2024samurai} &
64.1 & 73.1 & 72.0 & 69.6 & 62.5 & 70.3 & 68.0 & 70.2 & 69.1 & 68.0 & 63.2 & 63.0 & 69.6 & 64.5 
\\

\textbf{HiM2SAM(Ours)} &
67.4 & 76.5 & 74.8 & 71.7 & 65.1 & 72.5 & 71.3 & 72.5 & 71.7 & 70.6 & 65.7 & 65.9 & 71.8 & 67.3 
\\
\midrule
\multirow{2}{*}{\textbf{}} & 
\multicolumn{14}{c}{\textbf{LaSOT$_\text{ext}$}} \\

\midrule
SAMURAI~\cite{yang2024samurai} &
61.1 & 73.2 & 75.8 & 55.2 & 45.9 & 56.6 & 59.6 & 43.4 & 57.3 & 52.3 & 48.0 & 46.5 & 70.4 & 49.2 \\

\textbf{HiM2SAM(Ours)} &
65.7 & 75.8 & 77.2 & 57.2 & 47.4 & 58.4 & 60.3 & 43.5 & 59.8 & 54.4 & 50.1 & 47.1 & 70.8 & 47.7 
\\
\midrule
Average Gains &
\cellcolor{dblue} \textbf{4.0} & 
\cellcolor{dblue!75} \textbf{3.0} & 
\cellcolor{dblue!50} \textbf{2.1} & 
\cellcolor{dblue!50} \textbf{2.1} & 
\cellcolor{dblue!50} \textbf{2.1} & 
\cellcolor{dblue!40} \textbf{2.0} & 
\cellcolor{dblue!40} \textbf{2.0} & 
\cellcolor{dblue!25} \textbf{1.2} & 
\cellcolor{dblue!70} \textbf{2.6} & 
\cellcolor{dblue!60} \textbf{2.4} & 
\cellcolor{dblue!55} \textbf{2.3} & 
\cellcolor{dblue!33} \textbf{1.8} & 
\cellcolor{dblue!30} \textbf{1.3} & 
\cellcolor{dblue!10} \textbf{0.7}\\
\bottomrule
\end{tabularx}
\end{table}

\subsection{Ablation Studies and Runtime Analysis}

To evaluate each design component in HiM2SAM, we conduct ablation experiments on the LaSOT dataset using the SAM2.1-Large model. Components examined include motion estimation: Kalman Filter (KF) and Point Tracker (PT), and memory mechanisms: short-term memory (SM) and long-term memory (LM). 

Results in Table~\ref{tab:abl} show that each module improves tracking over the SAM2 baseline, with the full combination yielding the best performance and only modest runtime overhead. Runtime is measured on a single NVIDIA A6000 GPU and Intel i7-9700K CPU. Latency is averaged over all frames from a representative LaSOT subset. Notably, using only PT achieves higher accuracy than KF but at a significant computational cost. In contrast, HiM2SAM’s hierarchical motion estimation combines KF’s efficiency and PT’s precision, minimizing runtime overhead. The last row reports our final result, which demonstrates the complementarity of the modules and preserves the real-time capability of the original SAM2 model, validating HiM2SAM’s design as a fast and accurate long-term tracker.

\begin{table}[t]
\centering
\caption{Ablation study of HiM2SAM-L on the LaSOT dataset. 
"KF" denotes the Kalman Filter, "PT" denotes the PointTracker, "SM" refers to the short-term memory module, and "LM" refers to the long-term memory module.}

\begin{tabularx}{\textwidth}{
XXXXXXXl}
\toprule
\textbf{KF} & \textbf{PT} & \textbf{SM} & \textbf{LM} & \textbf{AUC(\%)} & \textbf{P$_{norm}$(\%)} & \textbf{P(\%)} & \textbf{Latency(ms)} \\
\midrule
\textcolor{red}{\ding{55}} & \textcolor{red}{\ding{55}} & \textcolor{red}{\ding{55}} & \textcolor{red}{\ding{55}} & 68.32 & 76.16 & 73.59 & Baseline\\
\textcolor{softgreen}{\ding{51}} & \textcolor{red}{\ding{55}} & \textcolor{red}{\ding{55}} & \textcolor{red}{\ding{55}} & 70.81 & 78.87 & 76.47 & +1.24\\
\textcolor{red}{\ding{55}} & \textcolor{softgreen}{\ding{51}} & \textcolor{red}{\ding{55}} & \textcolor{red}{\ding{55}} & 73.73 & 81.99 & 79.34 & +165.95\\
\textcolor{red}{\ding{55}} & \textcolor{red}{\ding{55}} & \textcolor{softgreen}{\ding{51}} & \textcolor{red}{\ding{55}} & 72.67 & 80.67 & 78.23 & +<0.01\\
\textcolor{red}{\ding{55}} & \textcolor{red}{\ding{55}} & \textcolor{red}{\ding{55}} & \textcolor{softgreen}{\ding{51}} & 73.71 & 81.43 & 79.18 & +0.39\\
\textcolor{softgreen}{\ding{51}} & \textcolor{red}{\ding{55}} & \textcolor{softgreen}{\ding{51}} & \textcolor{red}{\ding{55}} & 74.23 & 82.69 & 80.21 &+1.25\\
\textcolor{softgreen}{\ding{51}} & \textcolor{softgreen}{\ding{51}} & \textcolor{softgreen}{\ding{51}} & \textcolor{red}{\ding{55}} & 74.85 & 83.27 & 80.90 & +3.27\\
\rowcolor{myblue} % Set background color for the last row
\textcolor{softgreen}{\ding{51}} & \textcolor{softgreen}{\ding{51}} & \textcolor{softgreen}{\ding{51}} & \textcolor{softgreen}{\ding{51}} &\textbf{75.09} & \textbf{83.16} &\textbf{81.04} & +3.68\\
\bottomrule
\label{tab:abl}

\end{tabularx}
\end{table}

\begin{figure}[t]
    \centering
\includegraphics[width=\linewidth]{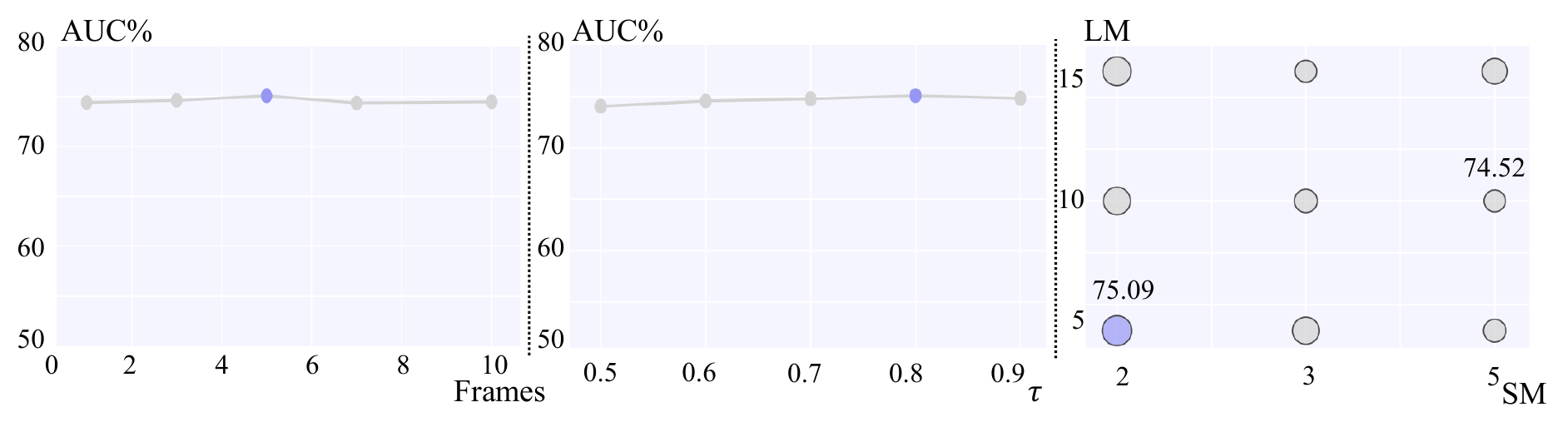}
    \caption{
        Sensitivity analysis of key parameters on the LaSOT dataset using HiM2sam-large.
        \textbf{Left}: AUC vs. the number of frames tracked by the point tracker.
        \textbf{Middle}: AUC vs. the IoU threshold $\tau$ in fine motion estimation.
        \textbf{Right}: Performance under different combinations of long-term memory (LM) and short-term memory (SM) frame intervals. Circle size indicates the corresponding AUC score.
    }
    \label{fig:param_sensitivity}
\end{figure}

\begin{figure}[!htbp]
    \centering  \includegraphics[width=1\linewidth]{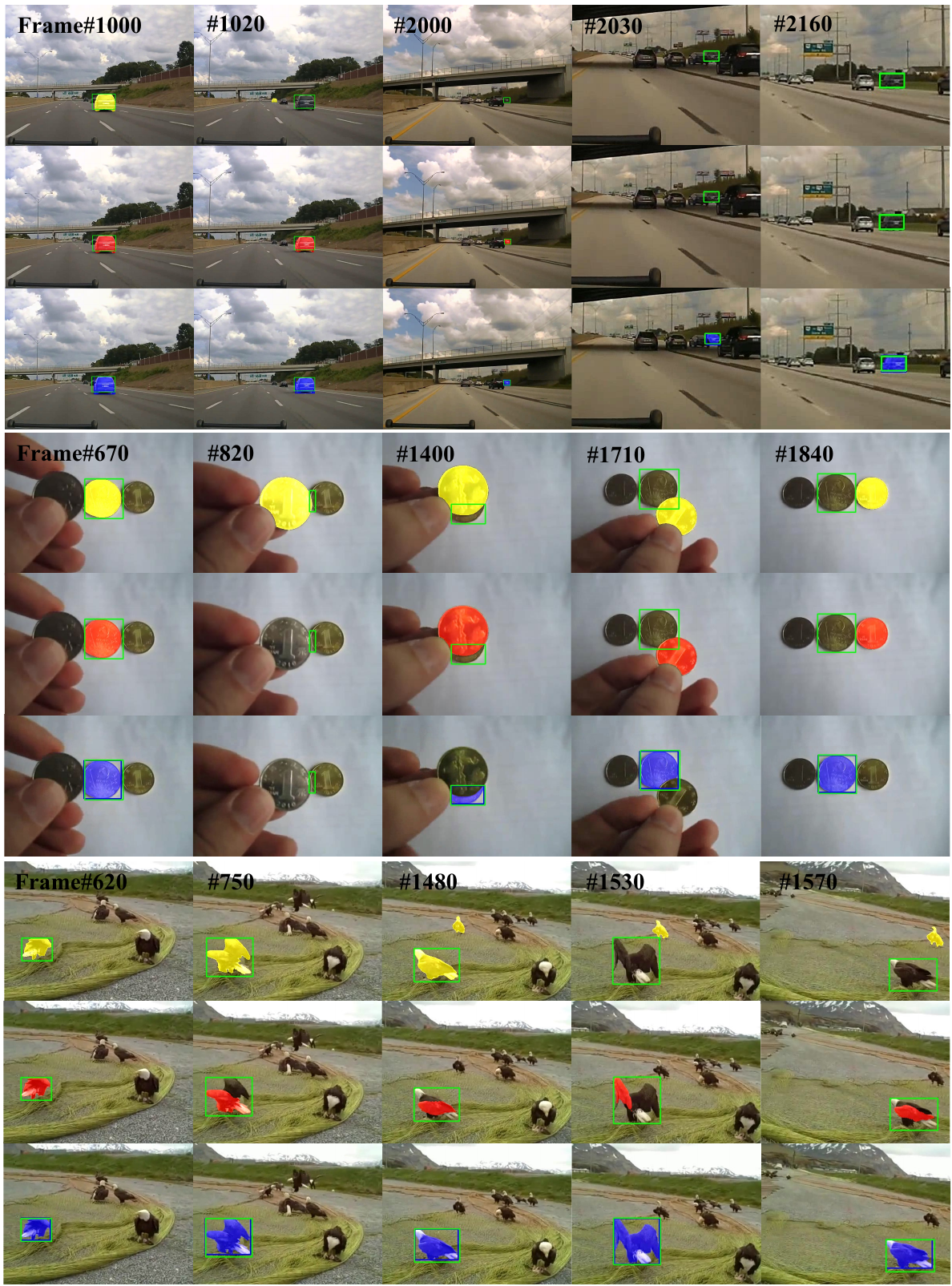}
    \caption{Qualitative comparison of tracking results on long video sequences.
     \textcolor{yellow}{Yellow} for SAMURAI, \textcolor{red}{red} masks represent DAM4SAM, and \textcolor{blue}{blue} for HiM2SAM (ours).
    \textcolor{green}{Green} bounding boxes denote ground truth.
    HiM2SAM provides more stable and accurate tracking results under long-term challenging scenarios.}

    \label{fig:quali}
\end{figure}

\subsection{ Sensitivity to Parameters }
We conduct a parameter sensitivity analysis on the LaSOT dataset to examine the robustness of our tracker HiM2SAM-L under different settings.
Figure~\ref{fig:param_sensitivity} presents the evaluation results for three key parameters:
(1) the number of frames used by the point tracker for temporal propagation;
(2) the IoU threshold $\tau$ used during coarse motion estimation to decide whether a fine motion estimation is needed (triggered when the IoU falls below $\tau$);
and (3) different combinations of frame intervals for long-term memory (LM) and short-term memory (SM).
We find that the tracker maintains overall stability across a wide range of parameter settings. Importantly, HiM2SAM consistently outperforms the SAM2 baseline, demonstrating the robustness and effectiveness of our method despite the presence of tunable parameters. The final selection of parameters is provided in our released code.

\subsection{ Qualitative Results}

We compare HiM2SAM with SAMURAI and DAM4SAM in challenging long-term tracking scenarios, as shown in Figure~\ref{fig:quali}. These sequences involve cluttered backgrounds, occlusions, and significant appearance variations. Our pixel-level motion estimation enables more precise mask boundaries, particularly under shape deformation. Meanwhile, the long-short memory mechanism enhances temporal consistency by improving robustness to occlusion and appearance changes. Together, these modules contribute to more stable and continuous tracking over long periods.

\section{Conclusion}
In this work, we enhance SAM2 for video object segmentation by introducing hierarchical motion estimation and memory optimization. Our approach combines lightweight and precise motion cues to handle complex dynamics. The split memory bank uses motion-aware short-term filtering and distractor-aware long-term selection to improve robustness against occlusions and appearance changes. Experiments on four long-term VOT benchmarks validate the effectiveness of our design in long-term and challenging tracking scenarios.

\FloatBarrier

\bibliographystyle{splncs04}
\small
\bibliography{LaTeX2e_Proceedings_Templates/ref}

\begin{thebibliography}{10}
\providecommand{\url}[1]{\texttt{#1}}
\providecommand{\urlprefix}{URL }
\providecommand{\doi}[1]{https://doi.org/#1}

\bibitem{rbf}
Buhmann, M.: Radial Basis Functions: Theory and Implementations. Cambridge University Press (2003)

\bibitem{cai2024hiptrackvisualtrackinghistorical}
Cai, W., Liu, Q., Wang, Y.: Hiptrack: Visual tracking with historical prompts (2024), \url{https://arxiv.org/abs/2311.02072}

\bibitem{cheng2023putting}
Cheng, H.K., Oh, S.W., Price, B., Lee, J.Y., Schwing, A.: Putting the object back into video object segmentation. In: arXiv (2023)

\bibitem{cheng2022xmem}
Cheng, H.K., Schwing, A.G.: {XMem}: Long-term video object segmentation with an atkinson-shiffrin memory model. In: European Conference on Computer Vision (ECCV) (2022)

\bibitem{cuttano2025samwise}
Cuttano, C., Trivigno, G., Rosi, G., Masone, C., Averta, G.: Samwise: Infusing wisdom in sam2 for text-driven video segmentation. In: Proceedings of the Computer Vision and Pattern Recognition Conference. pp. 3395--3405 (2025)

\bibitem{ding2024sam2long}
Ding, S., Qian, R., Dong, X., Zhang, P., Zang, Y., Cao, Y., Guo, Y., Lin, D., Wang, J.: Sam2long: Enhancing sam 2 for long video segmentation with a training-free memory tree. arXiv preprint arXiv:2410.16268  (2024)

\bibitem{fan2020lasothighqualitylargescalesingle}
Fan, H., Bai, H., Lin, L., Yang, F., Chu, P., Deng, G., Yu, S., Harshit, Huang, M., Liu, J., Xu, Y., Liao, C., Yuan, L., Ling, H.: Lasot: A high-quality large-scale single object tracking benchmark (2020), \url{https://arxiv.org/abs/2009.03465}

\bibitem{fan2019lasothighqualitybenchmarklargescale}
Fan, H., Lin, L., Yang, F., Chu, P., Deng, G., Yu, S., Bai, H., Xu, Y., Liao, C., Ling, H.: Lasot: A high-quality benchmark for large-scale single object tracking (2019), \url{https://arxiv.org/abs/1809.07845}

\bibitem{hsdf}
Huttenlocher, D., Klanderman, G., Rucklidge, W.: Comparing images using the hausdorff distance. IEEE Transactions on Pattern Analysis and Machine Intelligence  \textbf{15}(9),  850--863 (1993). \doi{10.1109/34.232073}

\bibitem{jiang2025sam2mot}
Jiang, J., Wang, Z., Zhao, M., Li, Y., Jiang, D.: Sam2mot: A novel paradigm of multi-object tracking by segmentation. arXiv preprint arXiv:2504.04519  (2025)

\bibitem{kf}
Kalman, R.E.: A new approach to linear filtering and prediction problems. Transactions of the ASME--Journal of Basic Engineering  \textbf{82}(Series D),  35--45 (1960)

\bibitem{kang2025exploring}
Kang, B., Chen, X., Lai, S., Liu, Y., Liu, Y., Wang, D.: Exploring enhanced contextual information for video-level object tracking. In: Proceedings of the AAAI Conference on Artificial Intelligence (2025)

\bibitem{karaev2024cotracker3}
Karaev, N., Makarov, I., Wang, J., Neverova, N., Vedaldi, A., Rupprecht, C.: {CoTracker3}: Simpler and better point tracking by pseudo-labelling real videos (2024)

\bibitem{karaev2023cotracker}
Karaev, N., Rocco, I., Graham, B., Neverova, N., Vedaldi, A., Rupprecht, C.: Cotracker: It is better to track together. In: European Conference on Computer Vision (ECCV) (2024)

\bibitem{kirillov2023segment}
Kirillov, A., Mintun, E., Ravi, N., Mao, H., Rolland, C., Gustafson, L., Xiao, T., Whitehead, S., Berg, A.C., Lo, W.Y., Dollár, P., Girshick, R.: Segment anything (2023), \url{https://arxiv.org/abs/2304.02643}

\bibitem{kristan2022tenth}
Kristan, M., Leonardis, A., Matas, J., Felsberg, M., Pflugfelder, R., K{\"a}m{\"a}r{\"a}inen, J.K., Chang, H.J., Danelljan, M., {\v{C}}ehovin~Zajc, L., Luke{\v{z}}i{\v{c}}, A., et~al.: The tenth visual object tracking vot2022 challenge results. In: European Conference on Computer Vision (ECCV). Springer (2022)

\bibitem{kristan2020eighth}
Kristan, M., Leonardis, A., Matas, J., Felsberg, M., Pflugfelder, R., K{\"a}m{\"a}r{\"a}inen, J.K., Danelljan, M., {\v{C}}ehovin~Zajc, L., Luke{\v{z}}i{\v{c}}, A., Drbohlav, O., et~al.: The eighth visual object tracking vot2020 challenge results. In: European Conference on Computer Vision (ECCV Workshops). Springer (2020)

\bibitem{li2025samjam}
Li, J., Cantu, F.J.P., Yu, E., Wong, A., Cui, Y., Chen, Y.: Samjam: Zero-shot video scene graph generation for egocentric kitchen videos. In: Proceedings of the Computer Vision and Pattern Recognition Conference. pp. 467--473 (2025)

\bibitem{lorat}
Lin, L., Fan, H., Zhang, Z., Wang, Y., Xu, Y., Ling, H.: Tracking meets lora: Faster training, larger model, stronger performance. In: European Conference on Computer Vision (ECCV) (2024)

\bibitem{dice}
Milletari, F., Navab, N., Ahmadi, S.A.: V-net: Fully convolutional neural networks for volumetric medical image segmentation (2016), \url{https://arxiv.org/abs/1606.04797}

\bibitem{fps}
Qi, C.R., Yi, L., Su, H., Guibas, L.J.: Pointnet++: Deep hierarchical feature learning on point sets in a metric space. arXiv preprint arXiv:1706.02413  (2017)

\bibitem{ravi2024sam2}
Ravi, N., Gabeur, V., Hu, Y.T., Hu, R., Ryali, C., Ma, T., Khedr, H., R{\"a}dle, R., Rolland, C., Gustafson, L., Mintun, E., Pan, J., Alwala, K.V., Carion, N., Wu, C.Y., Girshick, R., Doll{\'a}r, P., Feichtenhofer, C.: Sam 2: Segment anything in images and videos. arXiv preprint arXiv:2408.00714  (2024), \url{https://arxiv.org/abs/2408.00714}

\bibitem{ryali2023hiera}
Ryali, C., Hu, Y.T., Bolya, D., Wei, C., Fan, H., Huang, P.Y., Aggarwal, V., Chowdhury, A., Poursaeed, O., Hoffman, J., Malik, J., Li, Y., Feichtenhofer, C.: Hiera: A hierarchical vision transformer without the bells-and-whistles. In: Proceedings of the 40th International Conference on Machine Learning (ICML) (2023)

\bibitem{segu2024samba}
Segu, M., Piccinelli, L., Li, S., Yang, Y.H., Van~Gool, L., Schiele, B.: Samba: Synchronized set-of-sequences modeling for end-to-end multiple object tracking. arXiv preprint arXiv:2410.01806  (2024)

\bibitem{shi2024evptrack}
Shi, L., Zhong, B., Liang, Q., Li, N., Zhang, S., Li, X.: Explicit visual prompts for visual object tracking. In: Proceedings of the AAAI Conference on Artificial Intelligence (2024)

\bibitem{adpt}
Shim, K., Ko, K., Hwang, J., Kim, C.: Adaptrack: Adaptive thresholding-based matching for multi-object tracking. 2024 IEEE International Conference on Image Processing (ICIP) pp. 2222--2228 (2024)

\bibitem{boostrack}
Stanojevic, V.D., Todorovic, B.T.: Boosttrack: boosting the similarity measure and detection confidence for improved multiple object tracking. Machine Vision and Applications  \textbf{35}(3) (2024)

\bibitem{sun2022coarse}
Sun, G., Liu, Y., Ding, H., Probst, T., Van~Gool, L.: Coarse-to-fine feature mining for video semantic segmentation. In: Proceedings of the IEEE/CVF Conference on Computer Vision and Pattern Recognition (CVPR). pp. 3126--3137 (2022)

\bibitem{vaswani2023attentionneed}
Vaswani, A., Shazeer, N., Parmar, N., Uszkoreit, J., Jones, L., Gomez, A.N., Kaiser, L., Polosukhin, I.: Attention is all you need (2023), \url{https://arxiv.org/abs/1706.03762}

\bibitem{dam4sam}
Videnovic, J., Lukezic, A., Kristan, M.: A distractor-aware memory for visual object tracking with {SAM2}. In: Comp. Vis. Patt. Recognition (2025)

\bibitem{xie2024autoregressive}
Xie, J., Zhong, B., Mo, Z., Zhang, S., Shi, L., Song, S., Ji, R.: Autoregressive queries for adaptive tracking with spatio-temporal transformers. In: Proceedings of the IEEE/CVF Conference on Computer Vision and Pattern Recognition (CVPR). pp. 19300--19309 (2024)

\bibitem{yang2024samurai}
Yang, C.Y., Huang, H.W., Chai, W., Jiang, Z., Hwang, J.N.: Samurai: Adapting segment anything model for zero-shot visual tracking with motion-aware memory (2024), \url{https://arxiv.org/abs/2411.11922}

\bibitem{ucmc}
Yi, K., Luo, K., Luo, X., Huang, J., Wu, H., Hu, R., Hao, W.: Ucmctrack: Multi-object tracking with uniform camera motion compensation. Proceedings of the AAAI Conference on Artificial Intelligence  \textbf{38}(7),  6702--6710 (Mar 2024). \doi{10.1609/aaai.v38i7.28493}

\bibitem{zeng2021motr}
Zeng, F., Dong, B., Zhang, Y., Wang, T., Zhang, X., Wei, Y.: Motr: End-to-end multiple-object tracking with transformer. In: European Conference on Computer Vision (ECCV) (2022)

\bibitem{zheng2024odtrack}
Zheng, Y., Zhong, B., Liang, Q., Mo, Z., Zhang, S., Li, X.: Odtrack: Online dense temporal token learning for visual tracking. In: Proceedings of the AAAI Conference on Artificial Intelligence (2024)

\bibitem{zhou2025sam2}
Zhou, Y., Sun, G., Li, Y., Xie, G.S., Benini, L., Konukoglu, E.: When sam2 meets video camouflaged object segmentation: A comprehensive evaluation and adaptation. Visual Intelligence  \textbf{3}(1), ~10 (2025)

\bibitem{zhu2024clip}
Zhu, W., Cao, J., Xie, J., Yang, S., Pang, Y.: Clip-vis: Adapting clip for open-vocabulary video instance segmentation. IEEE Transactions on Circuits and Systems for Video Technology  (2024)

\end{thebibliography}



% LNCS Homepage, \url{http://www.springer.com/lncs}. Last accessed 4
\end{document}